\documentclass[10pt,twocolumn,letterpaper]{article}

\usepackage{iccv}
\usepackage{times}
\usepackage{epsfig}
\usepackage{graphicx}
\usepackage{amsmath}
\usepackage{amssymb}

\usepackage{caption}
\usepackage{color}
\usepackage{mathtools}

\DeclarePairedDelimiter\floor{\lfloor}{\rfloor}

\usepackage[pagebackref=true,breaklinks=true,letterpaper=true,colorlinks,bookmarks=false]{hyperref}
\usepackage{authblk}

\iccvfinalcopy 

\def\iccvPaperID{1773} 

\ificcvfinal\fi
\begin{document}

\title{Online Video Deblurring via Dynamic Temporal Blending Network}

\author[1]{Tae Hyun Kim}
\author[2]{Kyoung Mu Lee}
\author[1]{Bernhard Sch{\"o}lkopf}
\author[1]{Michael Hirsch}

\affil[1]{Department of Empirical Inference, Max Planck Institute for Intelligent Systems
}
\affil[2]{Department of ECE, ASRI, Seoul National University

\tt \small \{tkim,bernhard.schoelkopf,michael.hirsch\}@tuebingem.mpg.de, kyoungmu@snu.ac.kr
}

\renewcommand\Authands{ and }

\twocolumn[{%
	\renewcommand\twocolumn[1][]{#1}%
	\maketitle
	\begin{center}
		\centering
		\includegraphics[width=\textwidth]{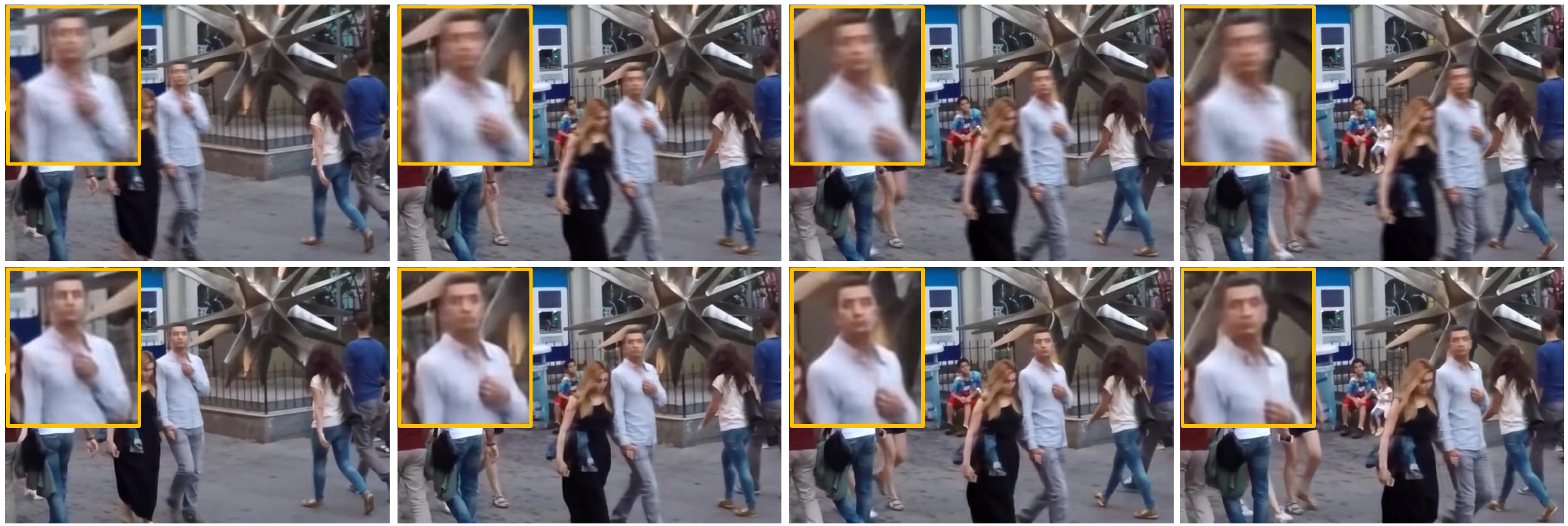}
		\captionof{figure}{Our online deblurring results (bottom) on a number of challenging real-world video frames (top) suffering from strong object motion. Our proposed approach is able to process the input video (VGA) in real-time, i.e. $\sim$24 fps on a standard graphics card (Nvidia GTX 1080).}
		\label{fig_intro}
	\end{center}%
	\vspace{3Ex}
}]

\maketitle

\begin{abstract}
\vspace{-2mm}
  State-of-the-art video deblurring methods are capable of removing
  non-uniform blur caused by unwanted camera shake and/or object
  motion in dynamic scenes. However, most existing methods are based
  on batch processing and thus need access to all recorded frames,
  rendering them computationally demanding and time consuming and thus
  limiting their practical use. In contrast, we propose an online
  (sequential) video deblurring method based on a spatio-temporal
  recurrent network that allows for real-time performance.
  In particular, we introduce a novel architecture which extends the
  receptive field while keeping the overall size of the network small
  to enable fast execution. In doing so, our network is able to remove
  even large blur caused by strong camera shake and/or fast moving
  objects. Furthermore, we propose a novel network layer that enforces
  temporal consistency between consecutive frames by dynamic temporal
  blending which compares and adaptively (at test time) shares features obtained at
  different time steps.  
  We show the superiority of the proposed
  method in an extensive experimental evaluation. 

\end{abstract}

\section{Introduction}

Moving objects in dynamic scenes as well as camera shake can cause undesirable motion blur in video recordings, often implying a severe degradation of video quality. This is especially true for low-light situations where the exposure time of each frame is increased, and for videos recorded with action (hand-held) cameras that have enjoyed widespread popularity in recent years. Therefore, not only to improve video quality~\cite{Cho2012video, Kim2015generalized} but also to facilitate other vision tasks such as tracking~\cite{Jin2005visual}, SLAM~\cite{Lee2011simultaneous}, and dense 3D reconstruction~\cite{Lee2013dense}, video deblurring techniques and their applications have seen an ever increasing interest recently. However, removing motion blur and restoring sharp frames in a blind manner (i.e., without knowing the blur of each frame) is a highly ill-posed problem and an active research topic in the field of computational photography. 

In this paper we propose a novel discriminative video deblurring method. Our method leverages recent insights within the field of deep learning and proposes a novel neural network architecture that enables run-times which are orders of magnitude faster than previous methods without significantly sacrificing restoration quality. Furthermore, our approach is the first online (sequential) video deblurring technique that is able to remove general motion blur stemming from both egomotion and object motion in real-time (for VGA video resolution).

Our novel network architecture employs deep convolutional residual networks \cite{He2016deep} with a layout that is recurrent both in time and space. For temporal sequence modeling we propose a network layer that implements a novel mechanism that we dub \emph{dynamic temporal blending}, which compares the feature representation at consecutive time steps and allows for dynamic (i.e. input-dependent) pixel-specific information propagation. Recurrence in the spatial domain is implemented through a novel network layout that is able to extend the spatial receptive field over time without increasing the size of the network. In doing so, we can handle large blurs better than typical networks for video frames, without run-time overhead.

Due to the lack of publicly available training data for video deblurring, we have collected a large number of blurry and sharp videos similar to the work of Kim et al.~\cite{Kim2016dynamic} and the recent work of Su et al.~\cite{Shuochen2017deep}. Specifically, we recorded sharp frames using a high-speed camera and generated realistic blurry frames by averaging over several consecutive sharp frames. Using this new dataset, we successfully trained our novel video deblurring network in an end-to-end manner. 

Using the proposed network and new dataset, we perform deblurring in a sequential manner, in contrast to many previous methods that require access to all frames, while at the same time being hundreds to thousands times faster than existing state-of-the-art video deblurring methods. In the experimental section, we validate the performance of our proposed model on a number of challenging real-world videos capturing dynamic scenes such as the one shown in Fig.~\ref{fig_intro}, and illustrate the superiority of our method in a comprehensive comparison with the state of the art, both qualitatively and quantitatively. 
In particular, we make the following contributions:
\begin{itemize}
	\setlength\itemsep{-1mm}
	\item we present, to the best of our knowledge, the first discriminative learning approach to video deblurring
	which is capable of removing spatially varying motion blurs in a sequential manner with real-time performance
	\item we introduce a novel spatio-temporal recurrent residual architecture with small computational footprint and increased receptive field along with a dynamic temporal blending mechanism that enables adaptive information propagation during test time
	\item we release a large-scale high-speed video dataset that enables discriminative learning
	\item we show promising results on a wide range of challenging real-world video sequences
\end{itemize}

\begin{figure*}[t]
	\begin{center}
		\includegraphics[width=\linewidth]{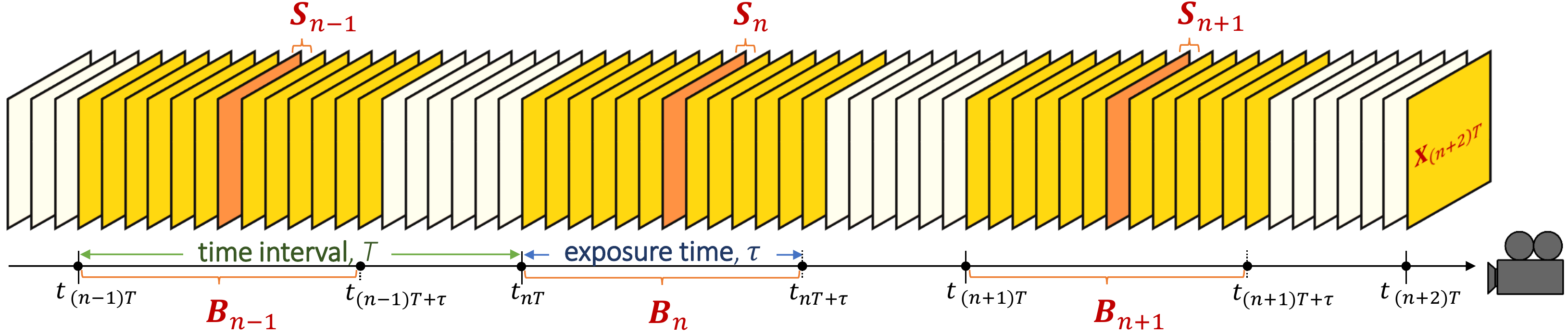}
	\end{center}
	\vspace{-3Ex}
	\caption{Generation of our blur dataset $\{\textbf{S}_n, \textbf{B}_n\}$ by averaging neighboring frames from a high-speed video $\{\textbf{X}_{nT}\}$.}
	\label{fig_dataset_generation}
		\vspace{-3Ex}
\end{figure*}

\section{Related Work}
\noindent{\bf{Multi-frame Deblurring.}} Early attempts to handle motion blur caused by camera shake considered multiple blurry images~\cite{Rav2005two,Chen2008robust}, and adapted techniques for removing uniform blur in single blurry images~\cite{Fergus2006removing,Shan2008high}. Other works include Cai et al~\cite{Cai2009blind}, and Zhang et al~\cite{Zhang2013multi} which obtained sharp frames by exploiting the sparsity of the blur kernels and gradient distribution of the latent frames. More recently, Delbracio and Sapiro~\cite{Delbracio2015burst} proposed Fourier Burst Accumulation (FBA) for burst deblurring, an efficient method to combine multiple blurry images without explicit kernel estimation by averaging complex pixel coefficients of multiple observations in the Fourier domain. Wieschollek et al.~\cite{Wieschollek2016burst} extended the work with a recent neural network approach for single image blind deconvolution \cite{Chakrabarti2016neural}, and achieved promising results by training the network in an end-to-end manner.

Most of the afore-mentioned methods assume stationarity, i.e., shift invariant blur, and cannot handle the more challenging case of spatially varying blur. To deal with spatially varying blur, often caused by rotational camera motion (roll) around the optical axis~\cite{Whyte2012non,Gupta2010single,Hirsch2011fast}, additional non-trivial alignment of multiple images is required. Several methods have been proposed to simultaneously solve the alignment and restoration problem~\cite{Cho2012registration,Zhang2014multi,Zhang2015intra}. In particular, Li et al.~\cite{Li2010generating} proposed a method to jointly perform camera motion (global motion) estimation and multi-frame deblurring, in contrast to previous methods that estimate a single latent image from multiple frames.

\noindent{\bf{Video Deblurring.}} Despite some of these methods being able to handle non-uniform blur caused by camera shake, none of them is able to remove spatially-varying blur stemming from object motion in a video recording of a dynamic scene. More generally, blur in a typical video might originate from various sources including moving objects, camera shake, depth variation, and thus it is required to estimate pixel-wise different blur kernels which is a highly intricate problem.

Some early approaches make use of sharp ``lucky'' frames which sometimes exist in long videos. Matsushita et al.~\cite{Matsushita2006full} detect sharp frames using image statistics, perform global image registration and transfer pixel intensities from neighboring sharp frames to blurry ones in order to remove blur.
Cho et al.~\cite{Cho2012video} improved deblurring quality significantly by employing additional local search and a blur model for aligning differently blurred image regions. However, their exemplar-based method still has some limitations in treating distinct blurs by fast moving objects due to the difficulty of accurately finding corresponding points between severely blurred objects.

Other deblurring attempts segment differently blurred regions. Both Levin \cite{levin2006blind} and Bar et al.~\cite{Bar2007variational} automatically segment a motion blurred object in the foreground from a (constant) background, and assume a uniform motion blur model in the foreground region.
Wulff and Black~\cite{Wulff2014modeling} consider differently blurred bi-layered scenes and estimate segment-wise accurate blur kernels by constraining those through a temporally consistent affine motion model. While they achieve impressive results especially at the motion boundaries, extending and generalizing their model to handle multi-layered scenes in real situations are difficult as we do not know the number and depth ordering of the layers in advance.

In contrast, there has been some recent work that estimates pixel-wise varying kernels directly without segmentation. Kim and Lee~\cite{Kim2014segmentation} proposed a method to parametrize pixel-wise varying kernels with motion flows in a single image, and they naturally extended it to deal with blurs in videos \cite{Kim2015generalized}. To handle spatially and temporally varying blurs, they parametrize the blur kernels with bi-directional optical flow between latent sharp frames that are estimated concurrently. Delbracio and Sapiro~\cite{delbracio2015hand} also use bi-directional optical flow for pixel-wise registration of consecutive frames, however manage to keep processing time low by using their fast FBA \cite{Delbracio2015burst} method for local blur removal.
Recently, Sellent et al.~\cite{Sellent2016stereo} tackled independent object motions with local homographies, and their adaptive boundary handling rendered promising results with stereo video datasets. 
Although these methods are applicable to remove general motion blurs, they are rather time consuming due to optical flow estimation and/or pixel-wise varying kernel estimation.
Probably closest related to our approach is the concurrent work of Su et al.~\cite{Shuochen2017deep}, which trains a CNN with skip connections to remove blur stemming from both ego and object motion. In a comprehensive comparison we show the merits of our novel network architecture both in terms of computation time as well as restoration quality.

\section{Training Datasets}
\label{sec:training}
A key factor for the recent success of deep learning in computer vision is the availability of large amounts of training data. However, the situation is more tricky for the task of blind deblurring. Previous learning-based single-image blind deconvolution \cite{Chakrabarti2016neural,Schuler2016learning,Sun2015learning} and burst deblurring \cite{Wieschollek2016burst} approaches have considered only ego motion and assumed a uniform blur model. However, adapting these techniques to the case of spatially and temporally varying motion blurs caused by both ego motion and object motion is not straightforward. Therefore, we pursue a different strategy and employ a recently proposed technique \cite{Kim2016dynamic,Shuochen2017deep,Nah2017deep} that generates pairs of sharp and blurry videos using a high-speed camera.

\begin{figure*}[t]
	\begin{center}
		\includegraphics[width=\linewidth]{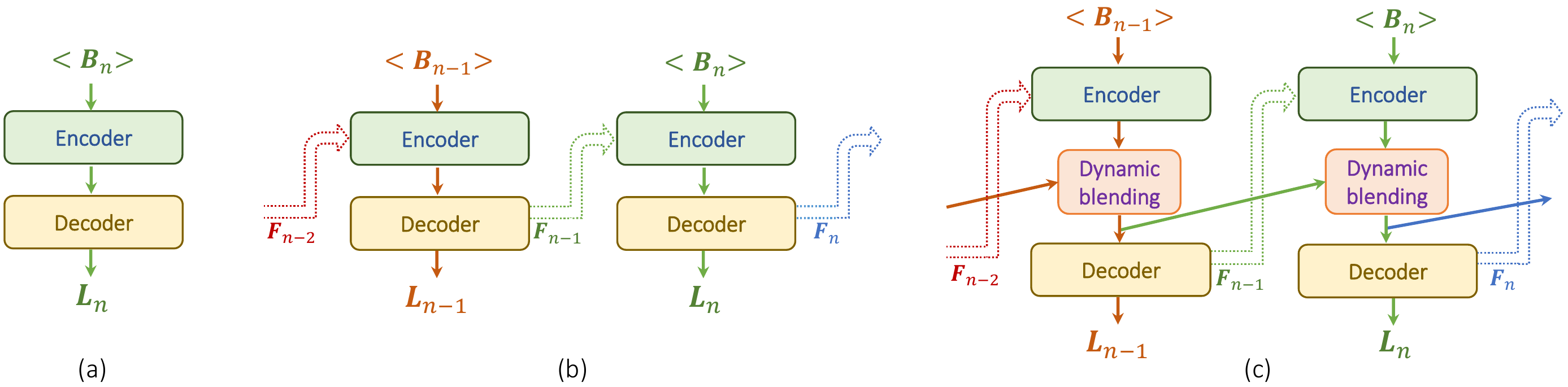}
	\end{center}
	\vspace{-4Ex}
	\caption{(a) Baseline model (CNN). (b) Spatio-temporal recurrent network (STRCNN). Feature maps at time step $n-1$ is added to the input of the network at time step $n$. (c) Spatio-temporal recurrent network with a proposed dynamic temporal blending layer (STRCNN+DTB). Intermediate feature maps are blended adaptively to render a clearer feature map by using weight map generated at runtime.}
	\label{fig_models}
	\vspace{-3Ex}
\end{figure*}

Given a high-speed video, we ``simulate'' long shutter times by averaging several consecutive short-exposure images, thereby synthesizing a video with fewer longer-exposed frames. The rendered (averaged) frames are likely to feature motion blur which might arise from camera shake and/or object motion. At the same time we use the center short-exposure image as a reference sharp frame. We thus have,
\begin{equation}
\begin{cases}
\textbf{B}_n = \frac{1}{\tau}\sum_{j=1}^{\tau} \textbf{X}_{nT+j}\\
\textbf{S}_n = \textbf{X}_{nT+\floor*{\frac{\tau}{2}}}
\end{cases},
\label{eq:datageneration}
\end{equation}
where $n$ denotes the time step, and $\textbf{X}_{nT}$, $\textbf{B}_{n}$, and $\textbf{S}_n$ are the short-exposure frame, synthesized blurry frame, and the reference sharp frame respectively. A parameter $\tau$ corresponds to the effective shutter speed which determines the number of frames to be averaged. A time interval, $\textit{T}$, which satisfies $\textit{T} \geq \tau$ controls the frame rate of the synthesized video. For example, the frame rate of the generated video is $\frac{\textit{f}}{\textit{T}}$ for a high-speed video captured at a frame rate  $\textit{f}$. Note that with these datasets, we can handle motion blurs only, but not other blurs (e.g., defocus blur).
We can control the strength of the blurs by adjusting $\tau$ (a larger $\tau$ generates more blurry videos), and can also change the duty cycle of the generated video by controlling the time interval $\textit{T}$. The whole process is visualized in Fig.~\ref{fig_dataset_generation}.

For our experiments, we collected high-speed sharp frames using a GoProHERO4 BLACK camera which supports recording HD (1280x720) video at a speed of $f=240$ frames per second, and then downsampled frames to the resolution of 960x540 size to reduce noise and jpeg artifacts. To generate more realistic blurry frames, we carefully captured videos to have small motions (ideally less than 1 pixel) among high-speed sharp frames as suggested in \cite{Kim2016dynamic}.
Moreover, we randomly selected parameters as $\tau \in \{7, 9, 11, 13, 15\}$ and $\tau \leq \textit{T} < 2\tau$ to generate various datasets with different blur sizes and duty cycles.

\section{Method Overview}\label{sec_method_overview}
In this paper, using our large dataset of blurry and sharp video pairs, we propose a video deblurring network estimating the latent sharp frames from blurry ones. As suggested in the work of Su et al.~\cite{Shuochen2017deep}, a straightforward and naive technique to deal with a video rather than a single image is employing a neural network repeatedly as shown in Fig.~\ref{fig_models} (a). Here, input to the network are consecutive blurry frames $\langle\textbf{B}_n\rangle_m=\{\textbf{B}_{n-m},\ldots, \textbf{B}_{n+m}\}$ where $\textbf{B}_n$ is the mid-frame and $m$ some small integer\footnote{For simplicity we dropped index $m$ from $\langle\textbf{B}_n\rangle_m$ in the figures.}. The network predicts a single sharp frame $\textbf{L}_n$ for time step $n$.
In contrast, we present networks specialized for treating videos by exploiting temporal information, and improve the deblurring performance drastically without increasing the number of parameters and overall size of the networks.

In the present section, we introduce network architectures which we have found to improve the performance significantly. First, in Fig.~\ref{fig_models}~(b), we propose a spatio-temporal recurrent network which effectively extends the receptive field without increasing the number of parameters of the network, facilitating the removal of large blurs caused by severe motion. Next, in Fig.~\ref{fig_models}~(c), we additionally introduce a network architecture that implements our dynamic temporal blending mechanism which enforces temporal coherence between consecutive frames and further improves our spatio-temporal recurrent model. In the following we describe our proposed network architectures in more detail.

\begin{figure*}[t]
	\begin{center}
		\includegraphics[width=\linewidth]{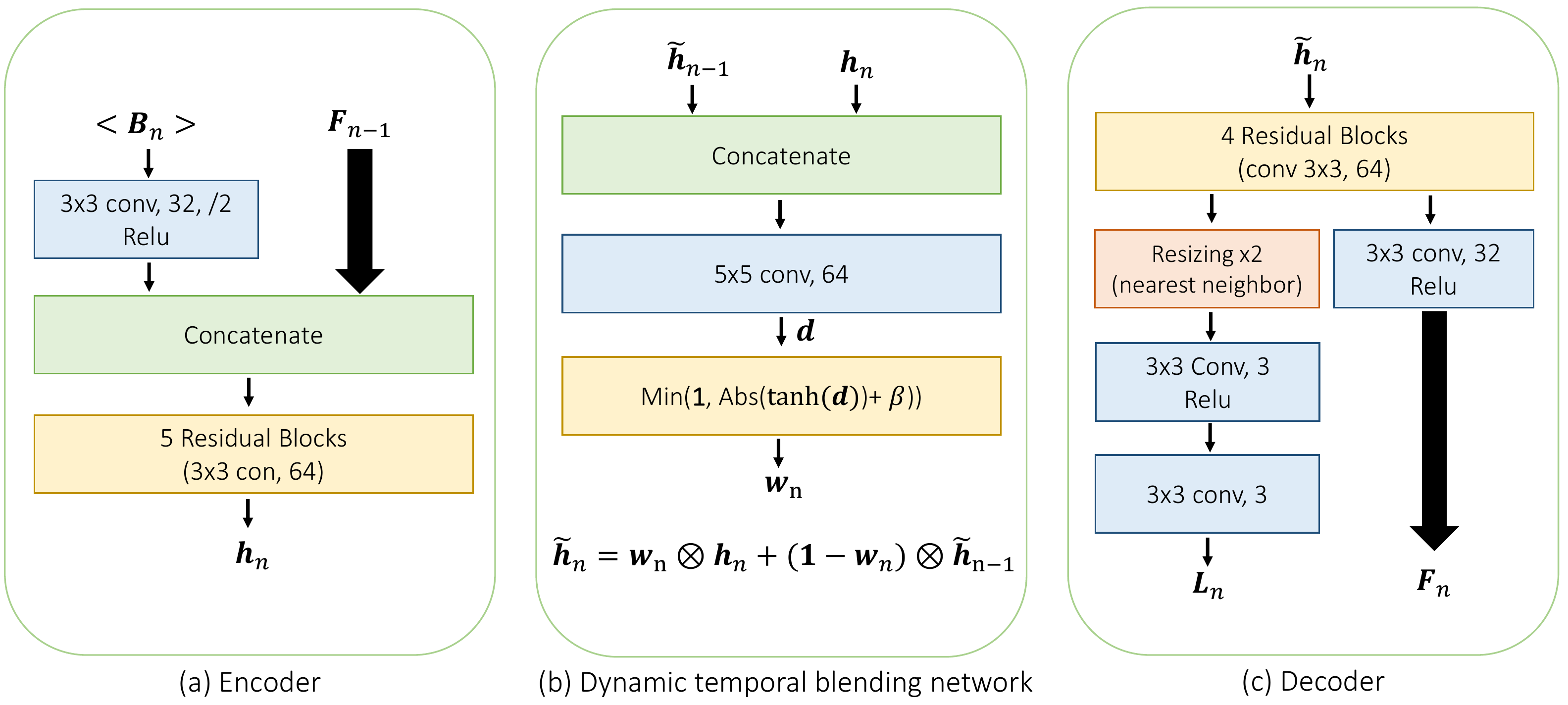}
	\end{center}
	\vspace{-4Ex}
	\caption{Detailed configurations of the proposed model. Our network is composed of encoder, dynamic temporal blending network, and decoder.}
	\label{fig_proposed_model}
	\vspace{-3Ex}
\end{figure*}

\begin{figure}[t]
	\begin{center}
		\includegraphics[width=0.6\linewidth]{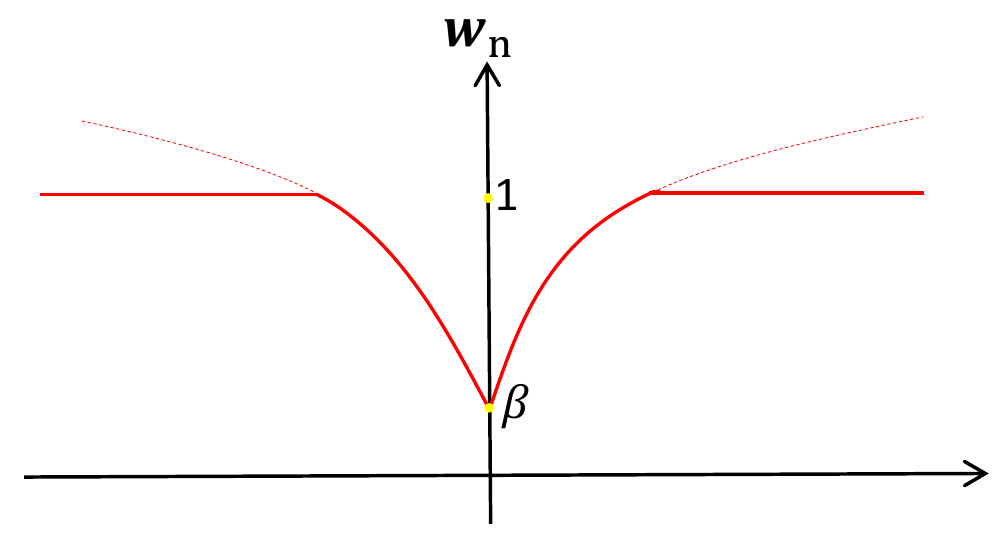}
	\end{center}
	\vspace{-4Ex}
	\caption{Sketch of weight parameter (weight map) $\textbf{w}_n$ at a location in 3D space, that resembles a traditional penalty function.}
	\label{fig_weight}
	\vspace{-2Ex}
\end{figure}

\subsection{Spatio-temporal recurrent network}
A large receptive field is essential for a neural network being capable of handling large blurs. For example, it requires about 50 convolutional layers to handle blur kernels of a size of 101x101~pixels with conventional deep residual networks using 3x3 small filters~\cite{He2016deep, He2016identity}. Although using a deeper network or larger filters are a straightforward and an easy way to ensure large receptive field, the overall run-time does increase with the number of additional layers and increasing filter size. Therefore, we propose an effective network which retains large receptive field without increasing its depth and filter size, i.e. number of layers and therewith its number of parameters.

The architecture of the proposed spatio-temporal network in Fig.~\ref{fig_models}~(b) is based on conventional recurrent networks~\cite{Wang2016saliency}, but has a point of distinction and profound difference. 
To be specific, we put $\textbf{F}_{n-1}$ which is the feature map of multiple blurry input frames $\langle\textbf{B}_{n-1}\rangle_m$ coupled with the previous feature map $\textbf{F}_{n-2}$ computed at time step $(n-1)$, as an additional input to our network together with blurry input frames $\langle\textbf{B}_n\rangle_m$ at time step $n$. By doing so, at time step $n$, the features of blurry frame $\textbf{B}_n$ passes through the same network ($m+1$) times, and ideally, we could increase the receptive field by the same factor without having to change the number of layers and parameters of our network. Notice that, in practice, the increase of receptive field is limited by the network capacity.

In other words, in a high dimensional feature space, each blurry input frame is recurrently processed multiple times by our recurrent network over time, thereby effectively experiencing a deeper spatial feature extraction with an increased receptive field. Moreover, further (temporal) information obtained from previous time steps is also  transferred to enhance the current frame, thus we call such a network \emph{spatio-temporal recurrent} or simply \emph{STRCNN}.

\subsection{Dynamic temporal blending network}
When handling video rather than single frames, it is important to enforce temporal consistency.
Although we recurrently transfer previous feature maps over time and implicitly share information between consecutive frames, 
we developed a novel mechanism for temporal information propagation that significantly improves the deblurring performance.

Motivated by the recent deep learning approaches of \cite{Brabandere2016dynamic,Jaderberg2015spatial}
which dynamically adapt network parameters to input data at test time, we also generate weight parameters for temporal feature blending that encourages temporal consistency, as depicted in Fig.~\ref{fig_models}~(c).
Specifically, based on our spatio-temporal recurrent network, we additionally propose a \textit{dynamic temporal blending} network, which generates weight parameter $\textbf{w}_n$ at time step $n$ which is used for linear blending between the feature maps of consecutive time steps, i.e.
\begin{equation}
\tilde{\textbf{h}}_n = \textbf{w}_n \otimes \textbf{h}_{n} + (\textbf{1} - \textbf{w}_n) \otimes \tilde{\textbf{h}}_{n-1},
\label{equ_temporal_filtering}
\end{equation}
where $\textbf{h}_{n}$ denotes the feature map at current time step $n$, $\tilde{\textbf{h}}_n$ denotes its filtered version, and $\tilde{\textbf{h}}_{n-1}$ denotes the previously filtered feature map at time step $(n-1)$. Weight parameters $\textbf{w}_n$ have a size equal to the size of $\textbf{h}_n$, and have values between zero and one. As a linear operator $\otimes$ denotes element-wise multiplication, our filter parameter $\textbf{w}_n$ can be viewed as a locally varying weight map.
Notably, $\textbf{h}_n$ is a feature activated in the middle of the entire network and thus it is different from $\textbf{F}_n$ which denotes the final activation.

It is natural that the previously filtered (clean) feature map $\tilde{\textbf{h}}_{n-1}$ is favored when ${\textbf{h}}_{n}$ is a degraded version of $\tilde{\textbf{h}}_{n-1}$. To do so, we introduce a new cell which generates filter parameter $\textbf{w}_n$ by comparing similarity between two feature maps, given by
\begin{equation}
\textbf{w}_n = \min(\textbf{1}, |\tanh(\textbf{A}\tilde{\textbf{h}}_{n-1} + \textbf{B}\textbf{h}_{n})| + \beta))
\end{equation}
where $\tanh(.)$ denotes a hyperbolic tangent function, $\textbf{A}$ and $\textbf{B}$ correspond to linear (convolutional) filters. A trainable parameter $0 \leq \beta \leq 1$ denotes a bias value, and it controls the mixing rate, i.e. it satisfies $\textbf{w}_n$ = $\beta$ when the hyperbolic tangent function returns zero.
As illustrated in Fig.~\ref{fig_weight}, $\textbf{w}_n$ denotes a typical penalty function at a location in 3-dimensional (feature) space, if $(\textbf{A}\tilde{\textbf{h}}_{n-1} + \textbf{B}\textbf{h}_{n})$ measures a proper distance between two feature maps.
 
Notably, to this end, we need only one additional convolutional layer and some non-linear activations such as $\tanh$, and thus, the computation is fast.
Although the proposed dynamic temporal blending network is simple and light, 
we demonstrate that it helps improve deblurring quality significantly in our experiments, and we refer to this network as \emph{STRCNN+DTB}.

\section{Implementation and Training}

In this section, we describe our proposed network architecture in full detail. An illustration is shown in Fig.~\ref{fig_proposed_model}, where we show a configuration at a single time step $n$ only since our model shares all trainable variables across time. Our network comprises three modules, i.e. encoder, dynamic temporal blending network, and decoder. Furthermore, we also discuss our objective function and training procedure.

\subsection{Network architecture}
\subsubsection{Encoder}
Figure~\ref{fig_proposed_model}~(a) depicts the \textit{encoder} of our proposed network. Input are $(2m+1)$ consecutive blurry frames $\langle\textbf{B}_n\rangle_m$ where $\textbf{B}_n$ is the mid-frame, along with feature activations $\textbf{F}_{n-1}$ from the previous stage.
All input images are in color and range in intensity from 0 to 1. The feature map $\textbf{F}_{n-1}$ is half the size of a single input image, and has 32 channels.
All blurry input images are filtered first, before being concatenated with the feature map and being fed into a deep residual network. Our encoder has a stack of 5 residual blocks (10 convolutional layers) similar to \cite{He2016deep}. Each convolution filter within a residual block is composed of 64 filters of size 3x3~pixels. The output of our encoder is feature map $\textbf{h}_{n}$.

\subsubsection{Dynamic temporal blending}

Our \textit{dynamic temporal blending} network is illustrated in Figure~\ref{fig_proposed_model}~(b). It takes two concatenated feature maps $\tilde{\textbf{h}}_{n-1}$ and $\textbf{h}_n$ as input and estimates weight maps $\textbf{w}_n$ through a convolutional layer with filters of size 5x5 pixels and a subsequent squashing function ($\tanh(.)$ and $Abs(.)$).
Finally, the generated weight map $\textbf{w}_n$ is used for blending between $\tilde{\textbf{h}}_{n-1}$ and $\textbf{h}_n$ according to Eq.~\ref{equ_temporal_filtering}.

We tested different layout configurations by changing the location of our dynamic temporal blending network. Best results were obtained when placing the dynamic temporal blending network right between encoder and decoder as shown in Fig.~\ref{fig_models}~(c) rather than somewhere in the middle of the encoder or decoder network.

\subsubsection{Decoder}
 Input to our \textit{decoder}, depicted in Fig.~\ref{fig_proposed_model} (c), is the blended feature map $\tilde{\textbf{h}}_{n}$ of the previous stage which is fed into a stack of 4 residual blocks (8 convolutional layers) with 64 convolutional filters of size 3x3~pixels. Outputs are a latent sharp frame $\textbf{L}_n$ that corresponds to the blurry input frame $\textbf{B}_n$, and a feature map $\textbf{F}_{n}$. As suggested in~\cite{Sajjadi2016Enhance}, we apply nearest neighbor upsampling to render the predicted output image being of the same size than the input frames. Notably, our output feature map $\textbf{F}_n$ is handed over as input to the network at the next time step.

\subsection{Objective function}
As an objective function we use the mean squared error (MSE) between the latent frames and their corresponding sharp ground-truth frames, i.e.

\begin{equation}
\textbf{E}_{mse} = \frac{1}{N_{mse}}\sum_n ||\textbf{S}_n - \textbf{L}_n||^2,
\end{equation}
where $N_{mse}$ denotes the number of pixels in a latent frame.
In addition, we use weight decay to prevent overfitting, i.e.
\begin{equation}
\textbf{E}_{reg} = \|\textbf{\textbf{W}}\|^2,
\end{equation}
where $\textbf{W}$ denotes the trainable network parameters.
Our final objective function  $\textbf{E}$ is given by
\begin{equation}
\textbf{E} = \textbf{E}_{mse} + \lambda \,\textbf{E}_{reg},
\end{equation}
where $\lambda$ trades off the data fidelity and regularization term. In all our experiments we set $\lambda$ to $10^{-5}$.

\subsection{Training parameters}
For training, we randomly select 13 consecutive blurry frames from artifically blurred videos (i.e., $\textbf{B}_1,\ldots,\textbf{B}_{13}$) , 
and crop a patch per frame. Each patch is 128x128~pixels in size, and a randomly chosen pixel location is used for cropping all 13 patches.
Moreover, we use a batch size of 8, and employ Adam~\cite{Kingma2014adam} for optimization with an initial learning rate of 0.0001, which is decreased exponentially (decay rate = 0.96) with an increasing number of iterations. 

\section{Experiments}

\begin{figure}[t]
	\begin{center}
		\includegraphics[width=\linewidth]{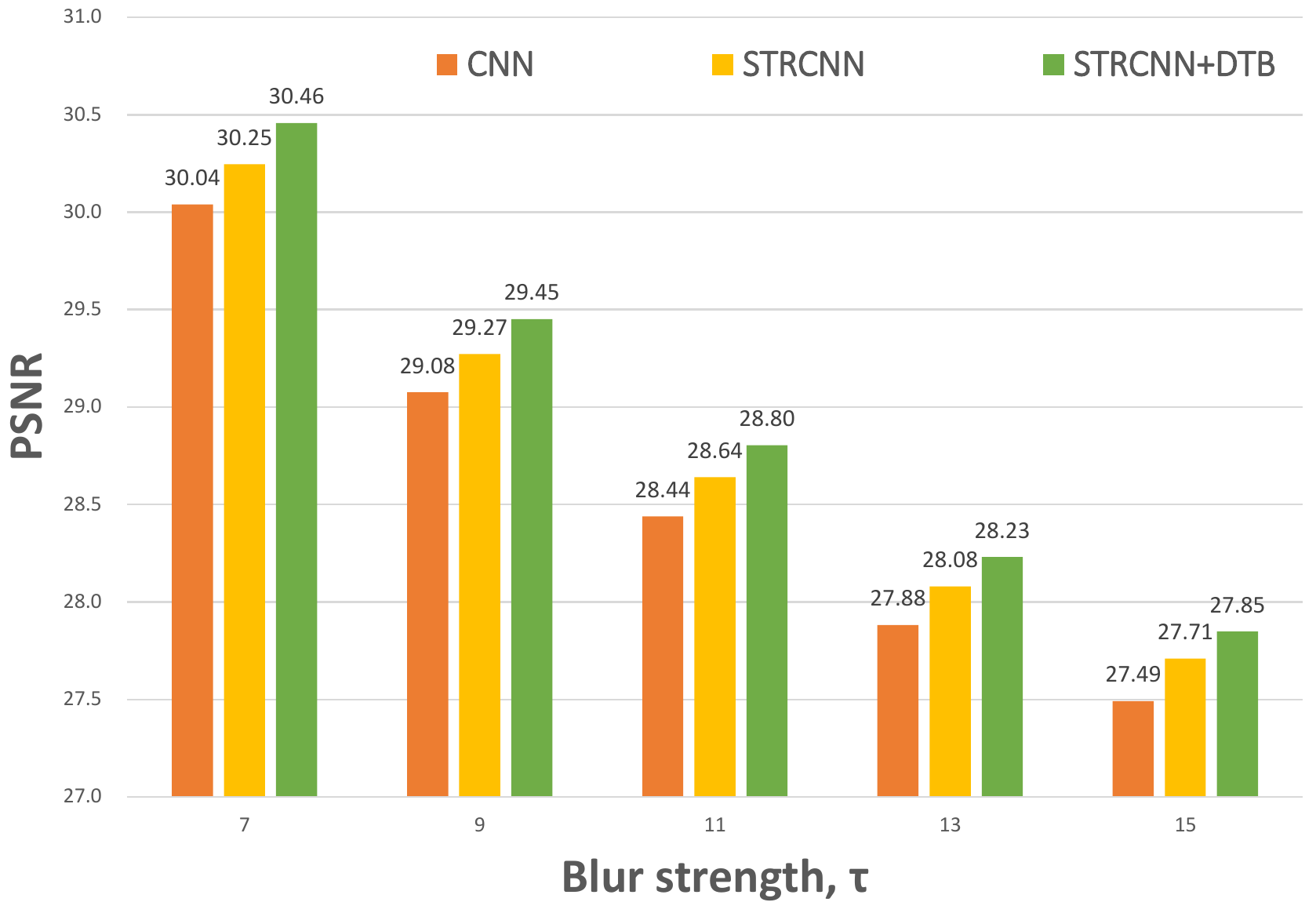}
	\end{center}
	\vspace{-4.5Ex}
	\caption{Performance comparisons among models proposed in Sec.~\ref{sec_method_overview} in terms of PSNR for varying blur strength.}
	\label{fig_comp_models}
	\vspace{-1.5Ex}
	
	\begin{center}
		\includegraphics[width=\linewidth]{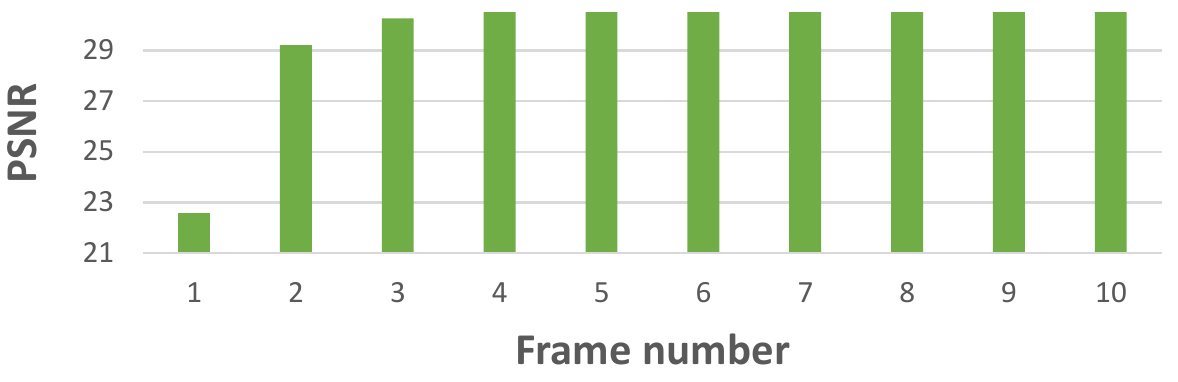}
	\end{center}
	\vspace{-4.5Ex}
	\caption{Average PSNR changes as frame increases.}
	\label{fig_framewise_performance}
	\vspace{-3Ex}
\end{figure}

\subsection{Model comparison}
We study the three different network architectures that we discussed in Sec.~\ref{sec_method_overview}, and evaluate deblurring quality in terms of peak signal-to-noise ratio (PSNR). For fair comparison, we use the same number of network parameters, except for one additional convolutional layer that is required in the dynamic temporal blending network. We use our own recorded dataset (described in Sec.~\ref{sec:training}) for training, and use the dataset of~\cite{Shuochen2017deep} for evaluation at test time.

First, we compare the PSNR values of the three different models for varying blur strength by changing the effective shutter speed $\tau$ in Eq.~(\ref{eq:datageneration}). We take five consecutive blurry frames as input to the networks.
As shown in Fig.~\ref{fig_comp_models}, our STRCNN+DTB model shows consistently better results for all blur sizes. On average, the PSNR value of our STRCNN is 0.2dB higher than the baseline (CNN) model, and STRCNN+DTB achieves a gain of 0.37dB against the baseline. 

Next, in Table~\ref{table_varying_inputs}, we evaluate and compare the performance of the models with a varying number of input blurry frames. Our STRCNN+DTB model outperforms other networks for all input settings. We choose STRCNN+DTB using five input frames ($m=2$) as our final model.

Our method processes a video sequence in an online fashion, thus we also show how the PSNR value changes with an increasing number of processed frames in Fig.~\ref{fig_framewise_performance}. Although our proposed method shows initially (i.e. n=1) worse performance due to lack of temporal information (initially zeros are given), restoration quality improves and stabilizes quickly after one or two frames.

\subsection{Quantitative results}
For objective evaluations, we compare with the state-of-the-art video deblurring methods~\cite{Kim2015generalized, Shuochen2017deep} whose source codes are available at the time of submission. In particular, as Shuochen et al.~\cite{Shuochen2017deep} provide their fully trained network parameters with three different input alignment methods.
Specifically, they align input images with optical flow (FLOW), or homography (HOMOG.), and they also take raw inputs without alignment (NOALIGN). For fair comparisons, we train our STRCNN+DTB model with their dataset, and evaluate performance with our own dataset.

We provide a quantitative comparison for 25 test videos captured with our high-speed camera described in Sec.\ref{sec:training}. Our model outperforms the state-of-the-art methods in terms of PSNR as shown in Table.~\ref{table_comparisons}.

\subsection{Qualitative results}
\begin{figure*}[t]
	\begin{center}
		\includegraphics[width=0.95\linewidth]{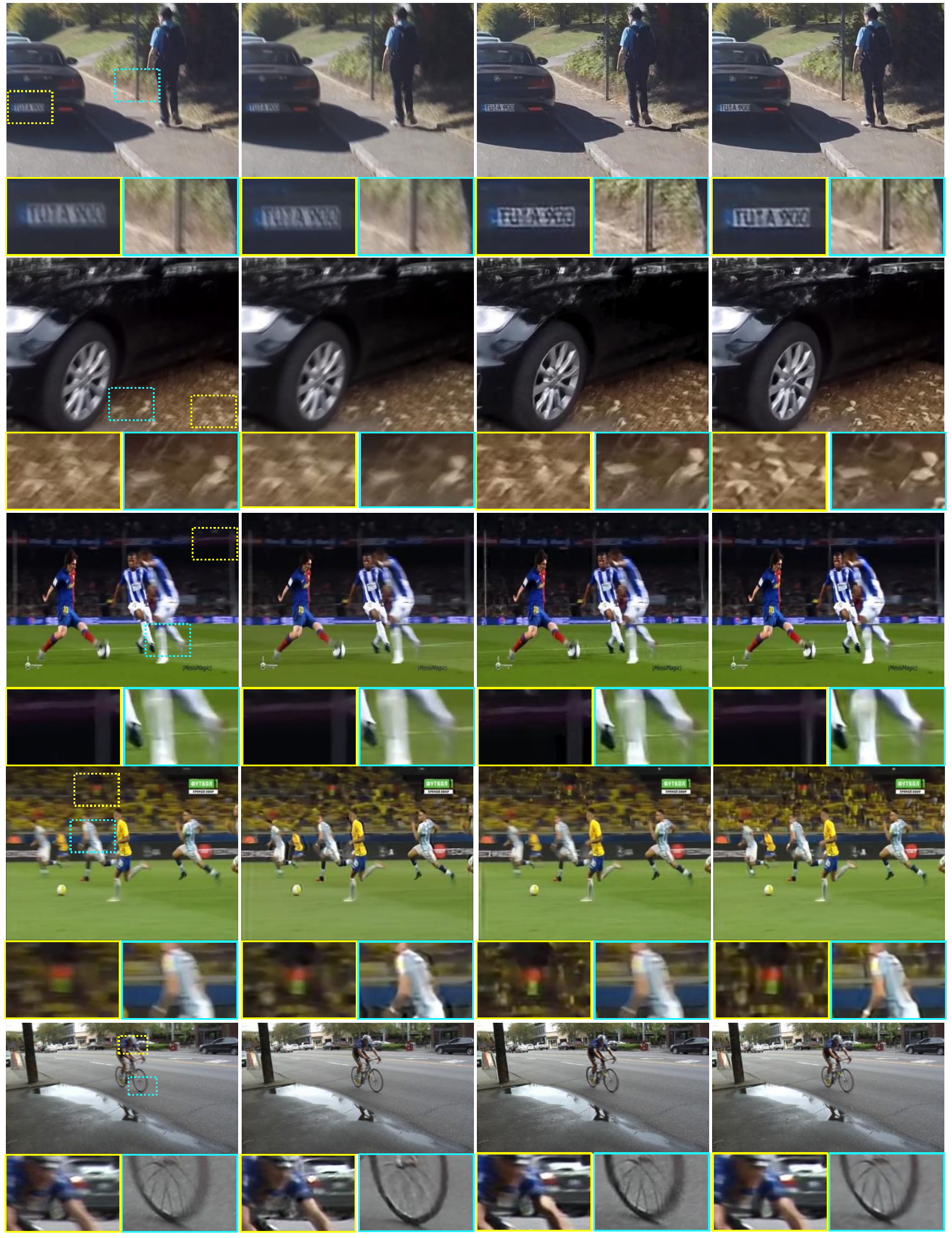}
	\end{center}
	\vspace{-4Ex}
	\caption{Left to right: Real blurry frames, Kim and Lee~\cite{Kim2015generalized}, Su et al.~\cite{Shuochen2017deep}, and our deblurring results.}
	\label{fig_comp_qualitative}
	\vspace{-2Ex}
	
\end{figure*}

To verify the generalization capabilities of our trained network, we provide qualitative results for a number of challenging videos.
Figure~\ref{fig_comp_qualitative} shows a comparison with \cite{Kim2015generalized, Shuochen2017deep} on challenging video clips. All these frames have spatially varying blurs caused by distinct object motion and/or rotational camera shake. In particular, blurry frames shown in the third and fourth rows are downloaded from YouTube, and thus contain high-level noise and severe encoding artifacts. Nevertheless, our method successfully restores the sharp frames especially at the motion boundaries in real-time.
In the last row, the offline (batch) deblurring approach by Kim and Lee \cite{Kim2015generalized} shows the best result however at the cost of long computation times. On the other hand, our approach yields competitive results though orders of magnitudes faster.

\begin{table}[]
	\centering
	
	\begin{tabular}{|c|c|c|c|}
		\hline
		\ Number of blurry inputs & 3     & 5     & 7     \\ \hline \hline			
		\ CNN & 29.20     & 29.25     & 29.03     \\ \hline 
		\ STRCNN & 29.45     & 29.63     & 29.37     \\ \hline 
		\ STRCNN+DTB      & 29.61 & \textbf{29.76} & 29.50 \\ \hline
	\end{tabular}
	\vspace{-1.5Ex}
	\caption{PSNR values for a varying number of input frames. STRCNN+DTB model, which encompasses a dynamic blending network, shows consistently better results.}
	\label{table_varying_inputs}
	\vspace{2Ex}

	\begin{tabular}{|c|c|c|c|}
		\hline
		Method &  \begin{tabular}[c]{@{}c@{}}PSNR \\ {(dB)}\end{tabular}   & \begin{tabular}[c]{@{}c@{}}Run-time\\ {(sec)}\end{tabular} & \begin{tabular}[c]{@{}c@{}}Online \\ mode\end{tabular} \\ \hline \hline
		Kim and Lee. \cite{Kim2015generalized} & 27.42 & $\sim$60000 (cpu) & x\\ \hline
		Cho et al. \cite{Cho2012video} & - & $\sim$6000 (cpu) & x \\ \hline
		Delbracio et al. \cite{delbracio2015hand} & - & $\sim$1500 (cpu) & o\\ \hline
		\begin{tabular}[c]{@{}c@{}}Su et al. \cite{Shuochen2017deep} \\ (FLOW)\end{tabular} & 28.81 & $\sim$564 (cpu+gpu) &o\\ \hline
		\begin{tabular}[c]{@{}c@{}}Su et al. \cite{Shuochen2017deep} \\ (HOMOG.)\end{tabular} & 28.09 & $\sim$154 (cpu+gpu) & o\\ \hline
		\begin{tabular}[c]{@{}c@{}}Su et al. \cite{Shuochen2017deep}\\ (NOALIGN)\end{tabular}   & 28.47 & $\sim$21 (gpu) & o \\ \hline
		STRCNN+DTB & 29.02 & $\sim$12.5 (gpu) & o\\ \hline
	\end{tabular}
	\vspace{-1Ex}
	\caption{Quantitative comparisons with state-of-the-art video deblurring method in terms of PSNR are given. A total of 25 (test) videos are used for evaluation. Moreover, execution times for processing 100 frames in HD resolution and comparison of online-processing capability are listed. }
	\label{table_comparisons}
	\vspace{-2Ex}
\end{table}

\subsection{Run time evaluations}
At test time, our online approach can process VGA (640x480) video frames at $\sim$24 frames per second with a recent NVIDIA GTX 1080 graphics card, and HD (1280x720) frames at $\sim$8 frames per second. In contrast, other conventional (offline) video deblurring methods take much longer. In Table.~\ref{table_comparisons}, we compare run-times for processing 100 HD (1280x720) video frames. 
Notably, our proposed method runs at a much faster rate than other conventional methods.

\subsection{Effects of dynamic temporal blending}
\begin{figure}[t]
	\begin{center}
          \rotatebox{90}{{\scriptsize \qquad STRCNN+DTB \hspace{13mm} STRCNN \hspace{17mm} Input}}
		\includegraphics[width=0.96\linewidth]{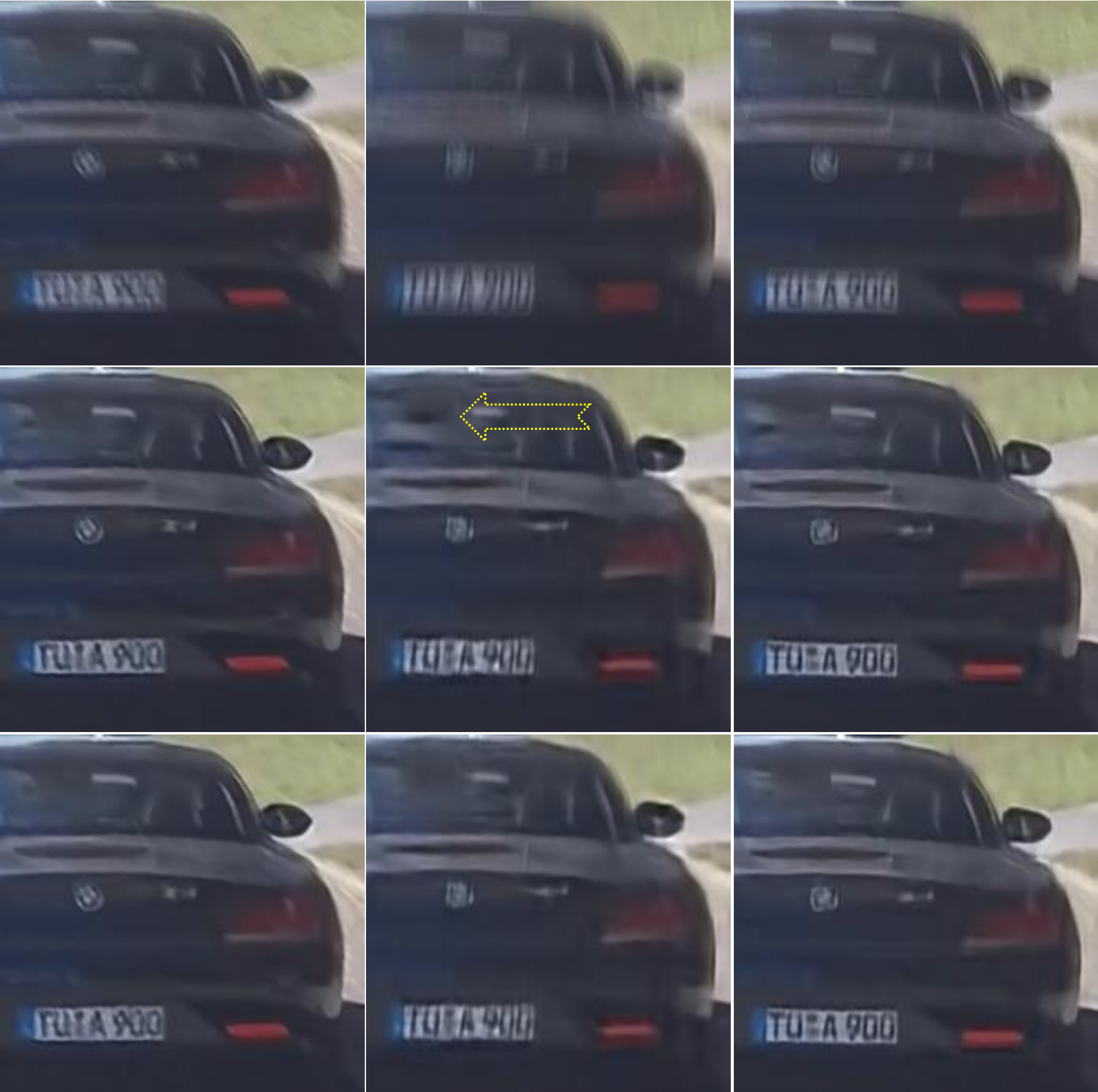}
	\end{center}
	\vspace{-4Ex}
	\caption{Top to bottom: Consecutive blurry frames, deblurred results by STRCNN and STRCNN+DTB. Yellow arrow indicates erroneous region caused by STRCNN model.}
	\label{fig_temporal_consistency}
	\vspace{-2Ex}
\end{figure}

In Fig.~\ref{fig_temporal_consistency}, we show a qualitative comparison of the results obtained with STRCNN and STRCNN+DTB. 
Although STRCNN could also remove motion blur by camera shake in the blurry frames well, it causes some artifacts on the car window. In contrast, STRCNN+DTB successfully restores sharp frames with less artifacts 
by enforcing temporal consistency using the proposed dynamic temporal blending network.

\section{Conclusion}
In this work, we proposed a novel network architecture for discriminative video deblurring.
To this end we have acquired a large dataset of blurry/sharp video pairs for training, and introduced a novel spatio-temporal recurrent network which enables near real-time performance by adding the feature activations of the last layer as an additional input to the network at the following time step. In doing so, we could retain large receptive field which is crucial to handle large blurs, without introducing a computational overhead. Furthermore, we proposed a dynamic blending network that enforces temporal consistency, which provides a significant performance gain. We demonstrate the efficiency and superiority of our proposed method by intensive experiments on challenging real-world videos.

{\small
\bibliographystyle{ieee}
\bibliography{ovd_iccv}

\begin{thebibliography}{10}\itemsep=-1pt

\bibitem{Bar2007variational}
L.~Bar, B.~Berkels, M.~Rumpf, and G.~Sapiro.
\newblock A variational framework for simultaneous motion estimation and
  restoration of motion-blurred video.
\newblock In {\em Proceedings of the IEEE International Conference on Computer
  Vision (ICCV)}, 2007.

\bibitem{Cai2009blind}
J.-F. Cai, H.~Ji, C.~Liu, and Z.~Shen.
\newblock Blind motion deblurring using multiple images.
\newblock {\em Journal of computational physics}, 228(14):5057--5071, 2009.

\bibitem{Chakrabarti2016neural}
A.~Chakrabarti.
\newblock A neural approach to blind motion deblurring.
\newblock In {\em Proceedings of the European Conference on Computer Vision
  (ECCV)}, 2016.

\bibitem{Chen2008robust}
J.~Chen, L.~Yuan, C.-K. Tang, and L.~Quan.
\newblock Robust dual motion deblurring.
\newblock In {\em Proceedings of the IEEE Conference on Computer Vision and
  Pattern Recognition (CVPR)}, 2008.

\bibitem{Cho2012registration}
S.~Cho, H.~Cho, Y.-W. Tai, and S.~Lee.
\newblock Registration based non-uniform motion deblurring.
\newblock {\em Computer Graphics Forum}, 31(7):2183--2192, 2012.

\bibitem{Cho2012video}
S.~Cho, J.~Wang, and S.~Lee.
\newblock Vdeo deblurring for hand-held cameras using patch-based synthesis.
\newblock {\em ACM Transactions on Graphics (SIGGRAPH)}, 2012.

\bibitem{Brabandere2016dynamic}
B.~De~Brabandere, X.~Jia, T.~Tuytelaars, and L.~Van~Gool.
\newblock Dynamic filter networks.
\newblock In {\em Advances in Neural Information Processing Systems (NIPS)},
  2016.

\bibitem{Delbracio2015burst}
M.~Delbracio and G.~Sapiro.
\newblock Burst deblurring: Removing camera shake through fourier burst
  accumulation.
\newblock In {\em Proceedings of the IEEE Conference on Computer Vision and
  Pattern Recognition (CVPR)}, 2015.

\bibitem{delbracio2015hand}
M.~Delbracio and G.~Sapiro.
\newblock Hand-held video deblurring via efficient fourier aggregation.
\newblock {\em IEEE Transactions on Computational Imaging}, 2015.

\bibitem{Fergus2006removing}
R.~Fergus, B.~Singh, A.~Hertzmann, S.~T. Roweis, and W.~T. Freeman.
\newblock Removing camera shake from a single photograph.
\newblock {\em ACM Transactions on Graphics (SIGGRAPH)}, 2006.

\bibitem{Gupta2010single}
A.~Gupta, N.~Joshi, C.~L. Zitnick, M.~Cohen, and B.~Curless.
\newblock Single image deblurring using motion density functions.
\newblock In {\em Proceedings of the European Conference on Computer Vision
  (ECCV)}, 2010.

\bibitem{He2016deep}
K.~He, X.~Zhang, S.~Ren, and J.~Sun.
\newblock Deep residual learning for image recognition.
\newblock In {\em Proceedings of the IEEE Conference on Computer Vision and
  Pattern Recognition (CVPR)}, 2016.

\bibitem{He2016identity}
K.~He, X.~Zhang, S.~Ren, and J.~Sun.
\newblock Identity mappings in deep residual networks.
\newblock In {\em Proceedings of the European Conference on Computer Vision
  (ECCV)}, 2016.

\bibitem{Hirsch2011fast}
M.~Hirsch, C.~J. Schuler, S.~Harmeling, and B.~Sch{\"o}lkopf.
\newblock Fast removal of non-uniform camera shake.
\newblock In {\em Proceedings of the IEEE International Conference on Computer
  Vision (ICCV)}, 2011.

\bibitem{Jaderberg2015spatial}
M.~Jaderberg, K.~Simonyan, A.~Zisserman, et~al.
\newblock Spatial transformer networks.
\newblock In {\em Advances in Neural Information Processing Systems (NIPS)},
  2015.

\bibitem{Jin2005visual}
H.~Jin, P.~Favaro, and R.~Cipolla.
\newblock Visual tracking in the presence of motion blur.
\newblock In {\em Proceedings of the IEEE Conference on Computer Vision and
  Pattern Recognition (CVPR)}, 2005.

\bibitem{Kim2014segmentation}
T.~H. Kim and K.~M. Lee.
\newblock Segmentation-free dynamic scene deblurring.
\newblock In {\em Proceedings of the IEEE Conference on Computer Vision and
  Pattern Recognition (CVPR)}, 2014.

\bibitem{Kim2015generalized}
T.~H. Kim and K.~M. Lee.
\newblock Generalized video deblurring for dynamic scenes.
\newblock In {\em Proceedings of the IEEE Conference on Computer Vision and
  Pattern Recognition (CVPR)}, 2015.

\bibitem{Kim2016dynamic}
T.~H. Kim, S.~Nah, and K.~M. Lee.
\newblock Dynamic scene deblurring using a locally adaptive linear blur model.
\newblock {\em arXiv preprint arXiv:1603.04265}, 2016.

\bibitem{Kingma2014adam}
D.~Kingma and J.~Ba.
\newblock Adam: A method for stochastic optimization.
\newblock {\em arXiv preprint arXiv:1412.6980}, 2014.

\bibitem{Lee2011simultaneous}
H.~S. Lee, J.~Kwon, and K.~M. Lee.
\newblock Simultaneous localization, mapping and deblurring.
\newblock In {\em Proceedings of the IEEE International Conference on Computer
  Vision (ICCV)}, 2011.

\bibitem{Lee2013dense}
H.~S. Lee and K.~M. Lee.
\newblock Dense 3d reconstruction from severely blurred images using a single
  moving camera.
\newblock In {\em Proceedings of the IEEE Conference on Computer Vision and
  Pattern Recognition (CVPR)}, 2013.

\bibitem{levin2006blind}
A.~Levin.
\newblock Blind motion deblurring using image statistics.
\newblock In {\em Advances in Neural Information Processing Systems (NIPS)},
  2006.

\bibitem{Li2010generating}
Y.~Li, S.~B. Kang, N.~Joshi, S.~M. Seitz, and D.~P. Huttenlocher.
\newblock Generating sharp panoramas from motion-blurred videos.
\newblock In {\em Proceedings of the IEEE Conference on Computer Vision and
  Pattern Recognition (CVPR)}, 2010.

\bibitem{Matsushita2006full}
Y.~Matsushita, E.~Ofek, W.~Ge, X.~Tang, and H.-Y. Shum.
\newblock Full-frame video stabilization with motion inpainting.
\newblock {\em IEEE Transactions on Pattern Analysis and Machine Intelligence
  (PAMI)}, 28(7):1150--1163, 2006.

\bibitem{Sajjadi2016Enhance}
M.~H. Mehdi SM~Sajjadi, Bernhard~Schölkopf.
\newblock Enhancenet: Single image super-resolution through automated texture
  synthesis.
\newblock {\em arXiv preprint arXiv:1612.07919}, 2016.

\bibitem{Nah2017deep}
S.~Nah, T.~H. Kim, and K.~M. Lee.
\newblock Deep multi-scale convolutional neural network for dynamic scene
  deblurring.
\newblock In {\em Proceedings of the IEEE Conference on Computer Vision and
  Pattern Recognition (CVPR)}, 2017.

\bibitem{Rav2005two}
A.~Rav-Acha and S.~Peleg.
\newblock Two motion-blurred images are better than one.
\newblock {\em Pattern recognition letters}, 26(3):311--317, 2005.

\bibitem{Schuler2016learning}
C.~J. Schuler, M.~Hirsch, S.~Harmeling, and B.~Schölkopf.
\newblock Learning to deblur.
\newblock {\em IEEE Transactions on Pattern Analysis and Machine Intelligence
  (PAMI)}, 38(7):1439--1451, 2016.

\bibitem{Sellent2016stereo}
A.~Sellent, C.~Rother, and S.~Roth.
\newblock Stereo video deblurring.
\newblock In {\em Proceedings of the European Conference on Computer Vision
  (ECCV)}, 2016.

\bibitem{Shan2008high}
Q.~Shan, J.~Jia, and A.~Agarwala.
\newblock High-quality motion deblurring from a single image.
\newblock {\em ACM Transactions on Graphics (SIGGRAPH)}, 2008.

\bibitem{Shuochen2017deep}
S.~Su, M.~Delbracio, J.~Wang, G.~Sapiro, W.~Heidrich, and O.~Wang.
\newblock Deep video deblurring.
\newblock In {\em Proceedings of the IEEE Conference on Computer Vision and
  Pattern Recognition (CVPR)}, 2017.

\bibitem{Sun2015learning}
J.~Sun, W.~Cao, Z.~Xu, and J.~Ponce.
\newblock Learning a convolutional neural network for non-uniform motion blur
  removal.
\newblock In {\em Proceedings of the IEEE Conference on Computer Vision and
  Pattern Recognition (CVPR)}, 2015.

\bibitem{Wang2016saliency}
L.~Wang, L.~Wang, H.~Lu, P.~Zhang, and X.~Ruan.
\newblock Saliency detection with recurrent fully convolutional networks.
\newblock In {\em Proceedings of the European Conference on Computer Vision
  (ECCV)}, 2016.

\bibitem{Whyte2012non}
O.~Whyte, J.~Sivic, A.~Zisserman, and J.~Ponce.
\newblock Non-uniform deblurring for shaken images.
\newblock {\em International Journal of Computer Vision}, 98(2):168--186, 2012.

\bibitem{Wieschollek2016burst}
P.~Wieschollek, B.~Sch{\"o}lkopf, H.~P. Lensch, and M.~Hirsch.
\newblock Burst deblurring: Removing camera shake through fourier burst
  accumulation.
\newblock In {\em Proceedings of the Asian Conference on Computer Vision
  (ACCV)}, 2016.

\bibitem{Wulff2014modeling}
J.~Wulff and M.~J. Black.
\newblock Modeling blurred video with layers.
\newblock In {\em Proceedings of the European Conference on Computer Vision
  (ECCV)}, 2014.

\bibitem{Zhang2014multi}
H.~Zhang and L.~Carin.
\newblock Multi-shot imaging: joint alignment, deblurring and
  resolution-enhancement.
\newblock In {\em Proceedings of the IEEE Conference on Computer Vision and
  Pattern Recognition (CVPR)}, 2014.

\bibitem{Zhang2013multi}
H.~Zhang, D.~Wipf, and Y.~Zhang.
\newblock Multi-image blind deblurring using a coupled adaptive sparse prior.
\newblock In {\em Proceedings of the IEEE Conference on Computer Vision and
  Pattern Recognition (CVPR)}, 2013.

\bibitem{Zhang2015intra}
H.~Zhang and J.~Yang.
\newblock Intra-frame deblurring by leveraging inter-frame camera motion.
\newblock In {\em Proceedings of the IEEE Conference on Computer Vision and
  Pattern Recognition (CVPR)}, 2015.

\end{thebibliography}
}

\end{document}